\documentclass[sigconf]{acmart}

\usepackage{booktabs} 
\usepackage{multirow}
\usepackage{wrapfig}
\usepackage{hyperref}
\usepackage{listings}
\usepackage{amsbsy}
\usepackage{color}
\usepackage{comment}
\usepackage{graphics}
\usepackage{graphicx}
\usepackage{amssymb}
\usepackage{amsmath} 
\usepackage{tabularx} 
\usepackage{multirow}
\usepackage{bbold}
\usepackage[utf8]{inputenc}
\usepackage[english]{babel}

\acmDOI{10.475/123_4}

\acmISBN{123-4567-24-567/08/06}

\acmConference[KDD'17]{the 23rd ACM SIGKDD International Conference on Knowledge Discovery and Data Mining}{Aug. 2017}{Halifax, CA} 
\acmYear{2017}
\copyrightyear{2017}

\newcommand\Mark[1]{\textsuperscript#1}

\begin{document}
\title{Adversary Resistant Deep Neural Networks with an Application to Malware Detection}

\author{Qinglong Wang\Mark{1}\Mark{2}, Wenbo Guo\Mark{1}, Kaixuan Zhang\Mark{1}, Alexander G. Ororbia II\Mark{1}, \\ Xinyu Xing\Mark{1}, C. Lee Giles\Mark{1}, Xue Liu\Mark{2}}
\affiliation{
  \institution{\Mark{1}Pennsylvania State University, \Mark{2}McGill University}
  \institution{}
}

\renewcommand{\shortauthors}{Q. Wang et al.}

\begin{abstract}

Beyond its highly publicized victories in Go, there have been numerous successful applications of deep learning in information retrieval, computer vision and speech recognition. In cybersecurity, an increasing number of companies have become excited about the potential of deep learning, and have started to use it for various security incidents, the most popular being malware detection. These companies assert that deep learning (DL) could help turn the tide in the battle against malware infections. However, deep neural networks (DNNs) are vulnerable to adversarial samples, a flaw that plagues most if not all statistical learning models. Recent research has demonstrated that those with malicious intent can easily circumvent deep learning-powered malware detection by exploiting this flaw.

In order to address this problem, previous work has developed various defense mechanisms that either augmenting training data or enhance model's complexity. However, after a thorough analysis of the fundamental flaw in DNNs, we discover that the effectiveness of current defenses is limited and, more importantly, cannot provide theoretical guarantees as to their robustness against adversarial sampled-based attacks. As such, we propose a new adversary resistant technique that obstructs attackers from constructing impactful adversarial samples by randomly nullifying features within samples. In this work, we evaluate our proposed technique against a real world dataset with 14,679 malware variants and 17,399 benign programs. We theoretically validate the robustness of our technique, and empirically show that our technique significantly boosts DNN robustness to adversarial samples while maintaining high accuracy in classification. To demonstrate the general applicability of our proposed method, we also conduct experiments using the MNIST and CIFAR-10 datasets, generally used in image recognition research.

\end{abstract}
\maketitle
\section{Introduction}
\label{sec:intro}

The past decades have witnessed the evolution of various malware detection technologies, ranging from signature-based solutions that compare an unidentified piece of code to known malware, to sandboxing solutions that execute a file within a virtual environment in order to determine whether the file is malicious or not. Unfortunately, none of these technologies seem to help much in the never-ending battle against malware infection. According to a recent report from Symantec Corporation~\cite{symantec}, one million malware variants hit the Internet every day that go completely undetected by many of the most common cybersecurity technologies in use today.

Substantial progress in neural network research, or deep learning (DL), has yielded a revolutionary approach to the cybersecurity community in the form of automatic feature learning. Recent research has demonstrated that malware detection approaches based on deep neural networks (DNNs) can recognize abstract complex patterns from a large amount of malware samples. This might offer a far better way to detect all types of malware, even in instances of heavy mutation~\cite{networkworld:deep-instinct,mittr:deep-instinct,bizety,wired:baidu,DBLP:SaxeB15,export:193768,blackhat:disassembly,Yuan:2014:DDL:2619239.2631434,mittrzeroday}. 

Despite its potential, deep neural architectures, like all other machine learning approaches, are vulnerable to what is known as adversarial samples ~\cite{Barreno:2010:SML:1860716.1860722,BiggioCMNSLGR13,Srndic:2014:PEL:2650286.2650798}. This means that these systems can be easily deceived by non-obvious and potentially dangerous manipulation~\cite{Szegedy2014Intriguing,goodfellow_explaining_2014,kdnuggetsfool,deepflawnews,kdnuggetsmisconception}. To be more specific, an adversary can infer the learning model underlying an application, examine feature/class importance, and identify the features that have greatest significant impact on correct classification. With this knowledge of feature importance, an adversary can, with minimal effort, craft an {\em adversarial sample} -- a synthetic example generated by slightly modifying a real example in order to make the deep learning system ``believe'' this modified sample belongs to an incorrect class with high confidence.

This flaw has been widely exploited to fool DNNs trained for image recognition (e.g.,~\cite{goodfellow_explaining_2014,Szegedy2014Intriguing,papernot2016limitations}). With the broad adoption of DNNs in malware detection, we speculate malware authors will also increasingly seek to exploit this vulnerability to circumvent malware detection. To note, recent research has already demonstrated that a malware author can leverage feature amplitude inequilibrium to bypass malware detectors powered by DNNs~\cite{grosse2016adversarial,deepdga}. 

Past research~\cite{goodfellow_explaining_2014, papernot2015distillation} in developing defense mechanisms relies on strong assumptions, which typically do not hold in many scenarios. Also, the techniques proposed can only be empirically validated and cannot provide any theoretical guarantees. This is particularly disconcerting when they are applied in a security critical application such as malware detection. 


In this work, we propose a new technical approach that can be empirically and theoretically guaranteed to be effective for malware detection and, more importantly, resistant to adversarial manipulation. To be specific, we introduce a random feature nullification in both the training and testing phase, which
makes a DNN model \emph{non-deterministic}. This non-deterministic nature manifests itself when attackers attempt to examine feature/class importance or when a DNN model takes input for classification. As such, there is a much lower probability
that attackers could correctly identify critical features for target DNN models. Even if attackers could infer critical features and construct a reasonable adversarial sample, the non-deterministic nature introduced in the model's processing of input will significantly reduce the effectiveness of these adversarial samples.


Technically speaking, our random feature nullification approach can also be viewed as stochastically ``dropping'' or omitting neuronal inputs. It can be viewed as a special case of dropout
regularization~\cite{srivastava2014dropout}, which involves randomly dropping unit activities (along with their connections), especially in the hidden layers, of a standard DNN. However, in normal drop-out, since a DNN is treated as deterministic at test-time~\footnote{In fact, ``inverted'' drop-out is applied in practice, which requires an extra division of the drop-out probability at training time in order to avoid the need for re-scaling at test-time. This specific implementation is used so that feed-forward inference is directly comparable to that under standard DNNs.}, critical features of the DNN model can still be correctly
identified and manipulated to create synthesized adversarial samples. Our approach is fundamentally different in that we nullify features at both train and test time. In Section~\ref{sec:eval}, we compare our random feature nullification with standard drop-out.


The simple approach proposed in this work is beneficial for a variety of reasons. First, it increases the difficulty for attackers to exploit the vulnerabilities of DNNs. Second, our adversary-resistant DNN maintains desirable classification performance while requiring only minimal modification to existing architectures. Third, the technique we propose can theoretically guarantee the resistance of DL to adversarial samples. Lastly, while this work is primarily motivated by the need to safeguard DNN models used in malware detection, it should be noted that the proposed technique is general and can be adopted to other applications where deep learning is popularly applied, such as image recognition. We demonstrate this applicability using two additional, publicly-available datasets in Section~\ref{sec:eval}.


The rest of the paper is organized as follows. In Section~\ref{sec:background}, we provide background on adversarial samples and, in Section~\ref{sec:rw}, we survey relevant work. Section~\ref{sec:tech} presents our technique and describes its properties. Experimental results appear in Section~\ref{sec:eval}, where we compare our technique to other approaches. Finally, conclusions are drawn in Section~\ref{sec:conclusion}.

\section{Background}
\label{sec:background}

Even though a well-trained model is capable of recognizing out-of-sample patterns, a deep neural architecture can be easily fooled by introducing perturbations to the input samples that are often indistinguishable to the human eye~\cite{Szegedy2014Intriguing}. These so-called ``blind spots'', or adversarial samples, exist because the input space of DNN is too broad to be fully explored~\cite{goodfellow_explaining_2014}. Given this, an adversary can uncover specific data samples in the input space and bypass DNN models. More specifically, it has been shown in~\cite{goodfellow_explaining_2014} that attackers can find the most powerful blind spots through effective optimization procedures. In multi-class classification tasks, the adversarial samples uncovered through this optimization can cause a DNN model to classify a data point into a class other than the correct one (and, oftentimes, not even a reasonable alternative).

Furthermore, according to~\cite{Szegedy2014Intriguing}, DNN models that share the same design goal, for example recognizing the same image set, all approximate a common highly complex, nonlinear function. Therefore, a relatively large fraction of adversarial examples generated from one trained DNN will be misclassified by other DNN models trained from the same data set but with different hyper-parameters. Given a target DNN, we refer to adversarial samples that are generated from other different DNN models but still maintain their attack efficacy against the target as \emph{cross-model adversarial samples}. 

Adversarial samples are generated by computing the derivative of the cost function with respect to the network's input variables. The gradient of any input sample represents a direction vector in this high-dimensional input space. Along this direction, any small change of this input sample will cause a DNN to generate a completely different prediction result. This particular direction is important since it represents the most effective way to degrade the performance of a DNN. Discovering this particular direction is done by passing the layer gradients from the output layer all the way back to the input layer via back-propagation. The gradient at the input may then be applied to the input samples to craft a adversarial examples.

To be more specific, let us define a cost function $\mathcal{L}(\theta,X,Y)$, where $\theta$, $X$ and $Y$ denotes the parameters of the DNN, the input dataset, and the corresponding labels respectively. In general, adversarial samples are created by adding an \emph{adversarial perturbation} $\delta X$ to real samples. In~\cite{goodfellow_explaining_2014}, the straightforward \emph{fast gradient sign} method was proposed for calculating adversarial perturbations as shown in~\eqref{eq:delta_X}:
\begin{equation}
  \begin{aligned}
  \label{eq:delta_X}
	\delta X = \phi \cdot sign(\mathcal{J}_{\mathcal{L}}(X)),
  \end{aligned}
\end{equation}
here $\delta X$ is calculated by multiplying the sign of the gradients of real sample $X$ with some coefficient $\phi$. $\mathcal{J}_{\mathcal{L}}(X)$ denotes the derivative of the cost function $\mathcal{L}(\cdot)$ with respect to $X$. $\phi$ controls the scale of the gradients to be added. 

An adversarial perturbation indicates the actual direction vector to be added to real samples. This vector drives a data point $X$ towards a direction that the cost function $\mathcal{L}(\cdot)$ is significantly sensitive to. However, it should be noted that $\delta X$ must be maintained within a small scale. Otherwise adding $\delta X$ will cause significant distortions to real samples, leaving the manipulation to be easily detected. 

\section{Related Work}
\label{sec:rw}

In order to counteract adversarial samples, recent research has been mainly focused on two different approaches -- data augmentation and model complexity enhancement. In this section, we summarize these techniques and discuss their limitations as follows.

\subsection{Data Augmentation}

To resolve the issue of ``blind spots'' (a more informal name given to adversarial samples), many methods that could be considered as sophisticated forms of data augmentation\footnote{Data augmentation refers to artificially expanding the data-set. In the case of images, this can involve deformations and transformations, such as rotation and scaling, of original samples to create new variants.} have been proposed (e.g.~\cite{goodfellow_explaining_2014,ororbia_ii_unifying_2016, gu2014towards}). In principle, these methods expand their training set by combining known samples with potential blind spots, the process of which has been called adversarial training~\cite{goodfellow_explaining_2014}. Here, we analyze the limitations of data augmentation mechanisms and argue that these limitations also apply to adversarial training methods.

Given the high dimensionality of data distributions that a DNN typically learns from, the input space is generally considered infinite~\cite{goodfellow_explaining_2014}. This implies that, for each DNN model, there could also be an adversarial space carrying an infinite amount of blind spots. Therefore, data augmentation based approaches face the challenge of covering these very large spaces. Since adversarial training is a form of data augmentation, such a tactic cannot possibly hope to cover an infinite space.

Adversarial training can be formally described as adding a regularization term known as \emph{DataGrad} to a DNN's training loss function~\cite{ororbia_ii_unifying_2016}. The regularization penalizes the directions uncovered by adversarial perturbations (introduced in Section~\ref{sec:background}). Therefore, adversarial training works to improve the worst case performance of a standard DNN. Treating the standard DNN much like a generative model, adversarial samples are produced via back-propagation and mixed into the training set and directly integrated into the model's learning phase. Despite the fact that there exists an infinite amount of adversarial samples, adversarial training has been shown to be effective in defending against those which are powerful and easily crafted. This is largely due to the fact that, in most adversarial training approaches~\cite{ororbia_ii_unifying_2016, goodfellow_explaining_2014}, adversarial samples can be generated efficiently for a particular type of DNN. The fast gradient sign method~\cite{goodfellow_explaining_2014} can generate a large pool of adversarial samples quickly while \emph{DataGrad}~\cite{ororbia_ii_unifying_2016} focuses on dynamically generating them per every parameter update. However, the simplicity and efficiency of generating adversarial samples also makes adversarial training vulnerable when these two properties are exploited to attack the adversarial training method \emph{itself}. Given that there exist infinite adversarial samples, we would need to repeat an adversarial training procedure each time a new adversarial example is discovered. Let us briefly consider DataGrad~\cite{ororbia_ii_unifying_2016}, which could be viewed as taking advantage of adversarial perturbations to better explore the underlying data manifold. While this leads to improved generalization, it does not offer any guarantees in covering all possible blind-spots. In this work, we do not address this issue by training a DNN model that covers the entire adversarial space. Rather, our design principle is to increase the difficulty for adversaries to find adversarial space in an efficient manner.

\subsection{Enhancing Model Complexity}

DNN models are already complex, with respect to both the nonlinear function that they try to approximate as well as their layered composition of many parameters. However, the underlying architecture is straightforward when it comes to facilitating the flow of information forwards and backwards, greatly alleviating the effort in generating adversarial samples. Therefore, several ideas~\cite{papernot2015distillation, gu2014towards} have been proposed to enhance the complexity of DNN models, aiming to improve the tolerance of complex DNN models with respect to adversarial samples generated from simple DNN models. 

\cite{papernot2015distillation} developed a \emph{defensive distillation} mechanism, which trains a DNN from data samples that are ``distilled'' from another DNN. By using the knowledge transferred from the other DNN, the learned DNN classifiers become less sensitive to adversarial samples. Although shown to be effective, this method is still vulnerable. This is because both DNN models used in this defense can be approximated by an adversary via training two other DNN models that share the same functionality and have similar performance. Once the two approximating DNN models are learned, the attacker can generate adversarial samples specific to this distillation-enhanced DNN model. Similar to~\cite{papernot2015distillation}, \cite{gu2014towards} proposed to stack an auto-encoder together with a standard DNN.  It shows that this auto-encoding enhancement increases a DNN's resistance to adversarial samples. However, the authors also admit that this stacked model can be easily approximated and exploited. 

Given the observation and analysis above, going beyond concealing adversarial space, we argue that an adversary-resistant DNN model also needs to be robust against adversarial samples generated from its best approximation. In light of these, this paper presents a new adversary-resistant DNN that not only increases the difficulty in finding its blind spots but also immunizes itself against adversarial samples generated from its best approximation.

\begin{figure}[t]
    \centering
    \begin{tabular}{c}
        \includegraphics[width=0.47\textwidth]{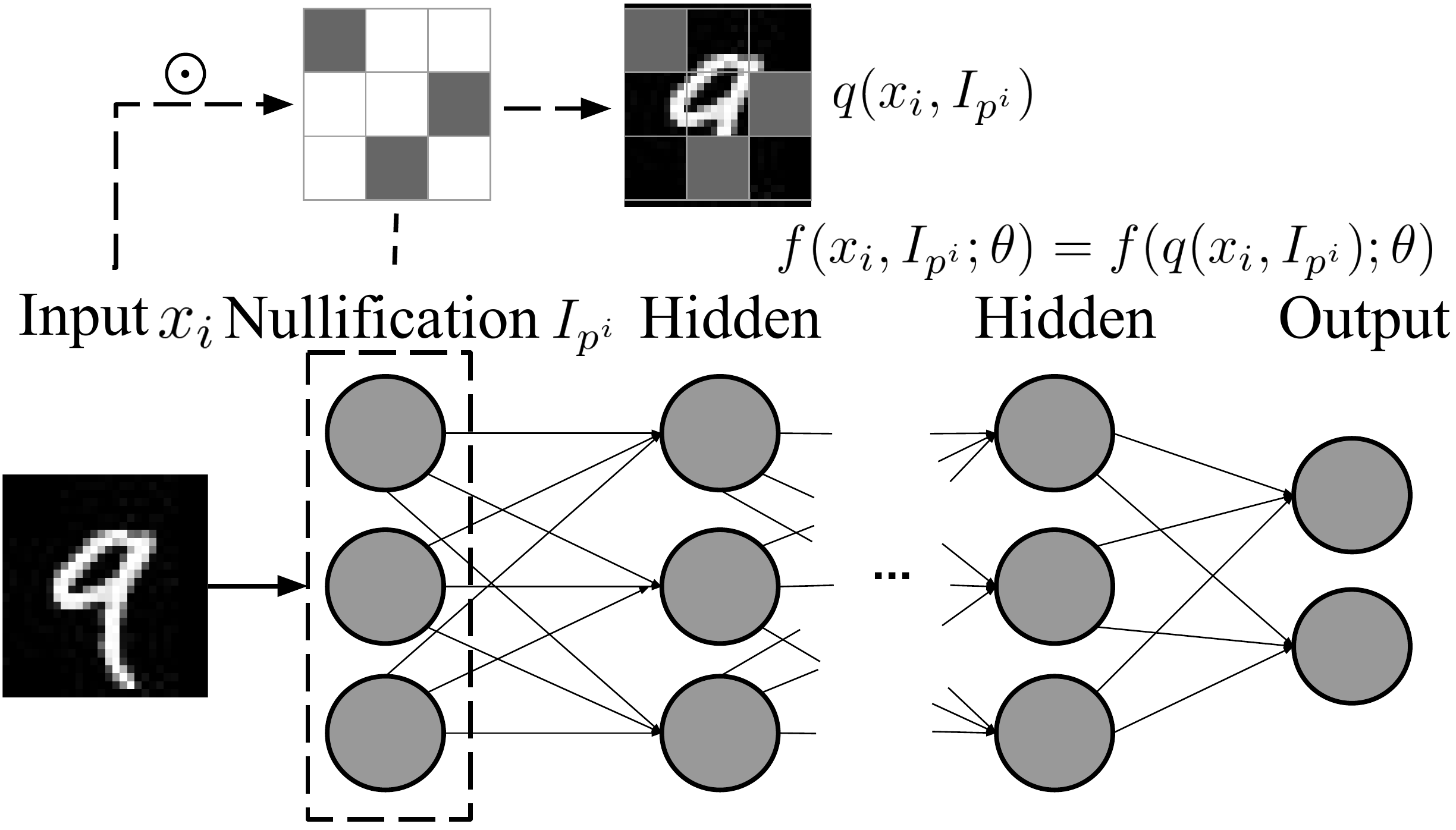}
    \end{tabular}
    \caption{A DNN equipped with a random feature nullification layer.}
    \vspace{-1em}
    \label{fig:rfn_dnn}
\end{figure}

\section{Random Feature Nullification}
\label{sec:tech}

Figure~\ref{fig:rfn_dnn} illustrates a DNN equipped with our random feature nullification method. Different from a standard DNN, it introduces an additional layer between the input and the first hidden layer. This intermediate layer is stochastic, serving as the source of randomness during both training and testing phases. In particular, it randomly nullifies or masks the features within the input. Let us consider image recognition as an example. When a DNN passes an image sample through the layer, it randomly cancels out some pixels within the image and then feeds the partially corrupted image to the first hidden layer. The proportion of pixels nullified is determined from hyper parameters $\mu_{p}$ and $\sigma_{p}^{2}$.

In this section, in addition to describing feature nullification and how to train a model using it, we will explain why our method offers theoretical guarantees of resistance and how it is different from other adversary-resistant techniques.

\subsection{Model Description}
\label{sec:dropout_input_tech}

Given input samples denoted by $X \in \mathbb{R}^{N\times M}$, where $N$ and $M$ denote the number of samples and features, respectively, random feature nullification amounts to simply performing element-wise multiplication of $X$ with $\hat{I_{p}}$. Here, $\hat{I_p} \in \mathbb{R}^{N\times M}$ is a mask matrix with the same dimensions as $X$. Note that in performing random nullification, it is inevitable that some feature information, which might be useful for classification, will be lost. To compensate for this, we choose to set a different nullification rate for each data sample. We hypothesize that this process could potentially lead to a better exploration of the input data's underlying manifold during training, ultimately obtaining a slightly better classification performance. We will verify this hypothesis with an experiment in malware classification in Section~\ref{sec:eval}. 

When training a DNN, for each input sample $x_i$ a corresponding $I_{p^{i}}$ is generated, where $I_{p^{i}}$ is a binary vector, where each element is either 0 or 1. In $I_{p^{i}}$, the total number of zeros, determined by $p^{i}$, are randomly distributed, following the uniform distribution. Formally, we denote the number of zeros in $I_{p^{i}}$ as $\lceil M\cdot p^{i}\rceil$, which will be randomly located, where $\lceil \cdot \rceil$ is the ceiling function. Note that $p^i$ is sampled from a Gaussian distribution $\mathrm{N}(\mu_{p},\sigma_{p}^{2})$. 


From Figure~\ref{fig:rfn_dnn}, random feature nullification can be viewed as a process in which a specialized layer simply passes nullified input to a standard DNN. As such, the objective function of a DNN with random feature nullification can be defined as follows.
\begin{equation}
  \begin{aligned}
  \label{eq:rndrop_dnn}
   \underset{\theta}{\text{min}}\sum_{i=1}^{N} \; \mathcal{L}\big (f(x_i, I_{p^i}; \theta), y_i \big ).
  \end{aligned}
\end{equation}
Here, $y_i$ is the label of the input $x_i$ and $\theta$ represents the set of model parameters. The random feature nullification process is represented by function $q(x_i, I_{p^i}) = x_i\odot I_{p^i}$, where $\odot$ denotes the \emph{Hadamard-Product} and $f(x_i, I_{p^i}; \theta)=f(q(x_i, I_{p^i}); \theta)$.

During training, Equation~\eqref{eq:rndrop_dnn} can be solved using stochastic gradient descent in a manner similar to that of a standard DNN. The only difference is that for each training sample, the randomly picked $I_{p^i}$ is fixed during forward and backward propagation until the next training sample arrives. This makes it feasible to compute the derivative of $\mathcal{L}\big (f(x_i, I_{p^i}; \theta), y_i \big )$ with respect to $\theta$ and update $\theta$ accordingly. During the testing process, when model parameters are fixed, in order to get stable test results, we use the expectation of the Gaussian distribution $\mathrm{N}(\mu_{p},\sigma_{p}^{2})$ as a substitute for the random variable $p^i$. More specifically, we generate a vector $I_p$ following the same procedure described earlier, but with $p$ equal to $\mu _{p}$.

\begin{figure}[t]
    \centering
    \begin{tabular}{c}
        \includegraphics[width=0.45\textwidth]{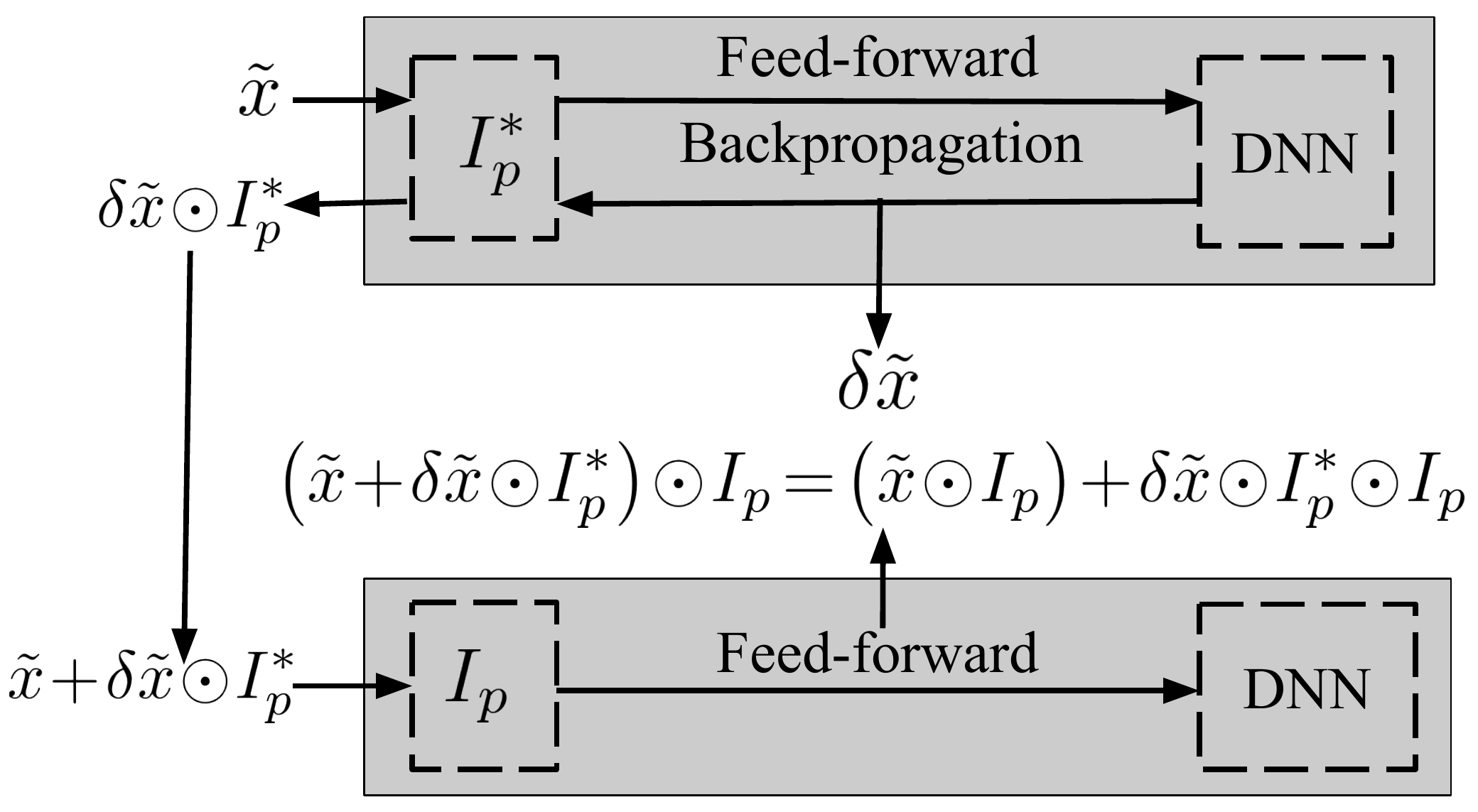}
    \end{tabular}
    \caption{An example of generating an adversarial sample and testing it on a DNN with random feature nullification.}
    \vspace{-1em}
    \label{fig:dnn_eq}
\end{figure}

\subsection{Analysis: Model Resistance to Adversaries}
\label{sec:analysis}

We will now present a theoretical analysis of our model's ability to resist adversarial samples. First, recall  (Section~\ref{sec:background}) that an adversary needs to generate adversarial perturbations in order to craft adversarial samples. According to Equation~\ref{eq:delta_X}, the adversarial perturbation is generated by computing the derivative of the DNN's cost function with respect to the input samples. 

Now let us assume that an adversary uses the same procedure to attack our proposed model. To be specific, the adversary computes the partial derivative of $\mathcal{L}\big (f(\tilde{x}, I_p; \theta), \tilde{y} \big )$ with respect to $\tilde{x}$, where $\tilde{x}$ denotes an arbitrary testing sample and $\tilde{y}$ denotes the corresponding label. More formally, the adversary needs to solve the following derivative:
\begin{equation}
  \begin{aligned}
  \label{eq:rndrop_adv_perturb}
	\mathcal{J}_{\mathcal{L}}(\tilde{x}) &= \frac{\partial \mathcal{L}\big(f(\tilde{x}, I_p; \theta), \tilde{y} \big)}{\partial \tilde{x}} \\
	&=  \mathcal{J}_{\mathcal{L}}(q) \cdot \frac{\partial q(\tilde{x}, I_p) }{\partial \tilde{x}}.
  \end{aligned}
\end{equation}
where $\mathcal{J}_{\mathcal{L}}(q) = \partial \mathcal{L}\big(f(\tilde{x}, I_p; \theta), \tilde{y} \big) /\partial q(\tilde{x}, I_p)$. Here, as mentioned earlier, $I_p$ is a mask matrix used during testing. Once the derivative above (Equation~\eqref{eq:rndrop_adv_perturb}) is calculated, an adversarial sample can be crafted by adding $\phi \cdot sign(\mathcal{J}_{\mathcal{L}}(\tilde{x}))$ to $\tilde{x}$, following~\cite{goodfellow_explaining_2014}.

To resolve Equation~\eqref{eq:rndrop_adv_perturb}, both $\mathcal{J}_{\mathcal{L}}(q)$ and $\partial q(\tilde{x}, I_p) / \partial \tilde{x}$ need to be computed. Note that $\mathcal{J}_{\mathcal{L}}(q)$ can be easily solved using back propagation of errors. However, term $\partial q(\tilde{x}, I_p) / \partial \tilde{x}$ carries random variable $I_p$, making  derivative computation impossible. In other words, it is the random variable $I_p$ itself that prohibits attackers from computing a derivative needed to produce an adversarial perturbation.

Recall that for each sample, the locations of zeroes within $I_p$ are randomly distributed. It is almost impossible for an adversary to pick up a value for $I_p$ that will match that which was randomly generated. Therefore, for this adversary, the best practice would be to approximate the value of $I_p$. To allow this adversary to make the best possible approximation, we further assume that the value of $p$ is known. With this assumption, one can randomly sample $I_p$ and treat it as a best approximation $I_p^*$. Using this approximation, the adversary then computes the most powerful adversarial perturbation. As shown in the top shaded region of Figure~\ref{fig:dnn_eq}, for the black-boxed DNN, we assume the most powerful adversarial perturbation is $\delta \tilde{x}$. Then, the adversarial perturbation for real sample $\tilde{x}$ is $\delta \tilde{x} \odot I_p^*$ . 

Assume the adversary uses a synthesized adversarial sample $\tilde{x} + \delta \tilde{x}  \odot I_p^*$ to attack the system shown in the bottom shaded region of Figure~\ref{fig:dnn_eq}. As we can see, the synthesized sample must pass through the the feature nullification layer before passing through the actual DNN. We describe this nullification in the following form.
\begin{equation}
  \begin{aligned}
  \label{eq:new_test}
	\big( \tilde{x} + \delta \tilde{x}  \odot I_p^* \big) \odot I_p = \big( \tilde{x} \odot I_p \big) + \delta \tilde{x} \odot I_p^* \odot I_p.
  \end{aligned}
\end{equation}
Here, $\tilde{x} \odot I_p$ is a nullified real sample, and $\delta \odot I_p^* \odot I_p$ represents the adversarial perturbation added to it. With $I_p^* \odot I_p$, even though $\delta \tilde{x}$ is the adversarial perturbation that impacts the DNN the most, this high-impact adversarial perturbation is still distorted and no longer represents the most effective perturbation for fooling the DNN. In Section~\ref{sec:eval}, we will provide empirical evidence to further validate this result.

In short, stochasticity, which naturally comes from $I_p$, is potentially our best defense against adversarial perturbation. It is also important to interpret our particular form of drop-out as a form of ``security through randomness''. Our parametrized feature nullification input layer, does not serve as a form of implicit model ensembling (or Bayesian averaging, which drop-out has been shown to be equivalent to in the case of single hidden-layer networks), especially given that randomness is still introduced at test-time.

\subsection{Comparison with Existing Defense Methods}
In the following, we thoroughly analyze the limited resistance provided by existing defense techniques introduced in Section~\ref{sec:rw}. According to~\cite{Kang2005,srivastava2014dropout,ororbia_ii_unifying_2016}, existing defense techniques can be generalized as training a standard DNN with various regularization terms (or even more generally as the DataGrad=regularized objective). More formally, the general objective is as follows: 
\begin{equation}
  \begin{aligned}
  \label{eq:dnn_reg}
	\underset{\theta}{\text{min}} \; \mathcal{G}(\theta, \tilde{x}, \tilde{y}) = \mathcal{L}(\theta, \tilde{x}, \tilde{y}) + \gamma \cdot \mathcal{R}(\theta, \tilde{x}, \tilde{y}),
  \end{aligned}
\end{equation}
where $\mathcal{L}(\theta, \tilde{x}, \tilde{y})$ is the training objective for a standard DNN, and $\mathcal{R}(\theta, \tilde{x}, \tilde{y})$ is a regularization term. Here, $\gamma$ controls the strength of the regularization. By adding regularization, \eqref{eq:dnn_reg} penalizes the direction represented by the adversarial perturbation that is optimal for crafting adversarial samples. 

However, existing defense methods that fall under this unifying framework are still vulnerable to adversarial samples problems, as shown below. To craft an adversarial sample from a model trained by solving~\eqref{eq:dnn_reg}, an adversary can easily produce an adversarial perturbation by computing the derivative with respect to a test sample $\tilde{x}$ as follows:
\begin{equation}
  \begin{aligned}
  \label{eq:adv_perturb}
	\mathcal{J}_{\mathcal{G}}(\tilde{x})  & = \frac{\partial \mathcal{G}(\theta, \tilde{x}, \tilde{y})}{\partial \tilde{x}} \\
	& = \frac{\partial \big ( \mathcal{L}(\theta, \tilde{x}, \tilde{y}) + \gamma \mathcal{R}(\theta, \tilde{x}, \tilde{y}) \big )}{\partial \tilde{x}}.
  \end{aligned}
\end{equation}

This indicates that prior studies only construct DNN models that are resistant to adversarial samples that target a standard DNN but do not build resistance to adversarial samples that would be generated to trick these newly ``hardened'' models. In addition, as we will show in Section~\ref{sec:eval}, the added regularization only imposes a limited penalty to the most effective adversarial perturbation.  Hence these methods might still be ineffective against adversarial samples that target standard DNNs, especially if an adversary simply increases the scale factor $\phi$ when generating adversarial samples. 

In other words, according to~\cite{goodfellow_explaining_2014}, the space containing both real samples and adversarial samples is too broad to be exhaustively explored. In the end, since adversarial training is a form of data augmentation, it cannot possibly hope to fully solve this problem. While all machine learning methods are susceptible to a broad space of adversarial samples, our proposed method, however, is a model-complexity-based approach that hardly adds any extra parameters, thus leaving the per-iteration run-time relatively untouched. 

\section{Evaluation} 
\label{sec:eval}

In this section, we first evaluate our proposed technique and compare it with adversarial training and dropout for a malware classification task using the dataset from \cite{berlin2015malicious}. Then we will show that our proposed method can be integrated with existing adversarial training methods and compare the combined approach's performance with both standalone methods -- random feature nullification (RFN) and adversarial training, respectively. Finally, we will demonstrate the generality of our proposed method by conducting some experiments in image recognition. In particular, we contrast our method with adversarial training and dropout on the MNIST \cite{lecun1998mnist} and CIFAR-10 \cite{krizhevsky2009learning} datasets.

\subsection{Datasets \& Experimental Design}
To comprehensively evaluate our method, we measure classification accuracy as well as model resistance to adversarial samples. In particular, to evaluate and compare the resistance of all three defense techniques, we test the DNN models against adversarial samples generated from the exact models trained either with RFN, adversarial training, and dropout. This means that we created three adversarial sample pools, one for each dataset (i.e., malware dataset, MNIST and CIFAR-10). The evaluation of resistance assumes that adversaries had acquired the full knowledge of each DNN model (i.e. hyper-parameters) and could construct the most effective adversarial samples to the best of their abilities. In this experimental setting, the observed resistance will then reflect a lower bound on model resistance against adversarial samples. For each dataset, we specify how to craft adversarial samples, especially with respect to the malware dataset.

\noindent{\textbf{Malware.}}
The malware dataset we experimented with is a collection of window audit logs\footnote{Window audit logs are collected using standard, built-in facilities, composed of two sources--users of a enterprise network as well as sandboxed virtual machine simulation runs using a set of malicious and benign binaries}. The dimensionality of the feature-space for each audit log sample is reduced to 10,000 according to the feature select metric used in \cite{berlin2015malicious}. Each feature indicates the occurrence of either a single event or a sequence of events\footnote{The number of events in one sequence can be as high as 3.}, thus taking on the value of 0 or 1. Here, 0 indicates that the sequence of events did not occur while 1 indicates the opposite. Classification labels are either 1, indicating a malware variant, or 0, indicating a benign program. The dataset is split into 26,078 training examples, with 14,399 benign software samples and 11,679 malicious software samples, and 6,000 testing samples, with 3,000 benign software samples and 3,000 malicious software samples. The task is to classify whether a given sample is benign or malicious. 

Adversarial perturbation for malware samples can be computed according to Equation \eqref{eq:delta_X}. However, a bit of care must be taken when generating adversarial samples for the malware dataset. Malware samples are usually represented by features that take on discrete and finite values, e.g. records of file system accesses, types of system calls incurred, etc. Therefore, it is more appropriate to use the $l_{0}$ distance:
\begin{equation}
  \begin{aligned}
  \label{eq:bound}
	||\hat{x} - x ||_{0} < \varepsilon ,
  \end{aligned}
\end{equation}
where $\hat{x} = x + \delta{x}$ represents adversarial samples generated from legitimate sample $x$. 

A similar approach was also adopted in~\cite{grosse2016adversarial}. Furthermore, as discussed in~\cite{grosse2016adversarial}, malware data contains stricter semantics in comparison to image data. In our case, each feature of a malware sample indicates whether or not a potential bit of malware has initiated a certain file system access. Therefore, large-scale manipulations across all features, as is typically done with image data, may break down a malicious program's functionality. To avoid this, we restrict the total number of manipulations that can occur per malware sample to be as small as possible. In this paper's setting, we set this to be 10. Moreover, since removing certain file system calls may also jeopardize a malware's internal logic, we further restrict the manipulation by only allowing the addition of new file system accesses. This equivocates to only positive manipulations, i.e. changing a feature from 0 to 1. Finally, since malware manipulation is done with the intent of fooling a DNN malware classifier, there is no need to modify a benign application such that it is classified as malicious. Therefore, in our experiments we only generate adversarial samples from the malware data points. In Table~\ref{tab:mal_advSample}, we show a few examples of features added to a malware sample. These added features only cause the malware to call several dynamically linked library files without damaging the program's malicious intent.

\begin{table}[t]
\small
\centering
\begin{tabular}{ll}
\hline
\multicolumn{2}{c}{\multirow{2}{*}{Examples of Changed Features}}                                                                                                                 \\
\multicolumn{2}{c}{}                                                                                                                                          \\ \hline
\multicolumn{2}{l}{\begin{tabular}[c]{@{}l@{}}WINDOWS\_FILE:Execute:{[}system{]}$\setminus$slc.dll, \\ WINDOWS\_FILE:Execute:{[}system{]}$\setminus$cryptsp.dll\end{tabular}}     \\ \hline
\multicolumn{2}{l}{\begin{tabular}[c]{@{}l@{}}WINDOWS\_FILE:Execute:{[}system{]}$\setminus$wersvc.dll, \\ WINDOWS\_FILE:Execute:{[}system{]}$\setminus$faultrep.dll\end{tabular}} \\ \hline
\multicolumn{2}{l}{\begin{tabular}[c]{@{}l@{}}WINDOWS\_FILE:Execute:{[}system{]}$\setminus$imm32.dll, \\ WINDOWS\_FILE:Execute:{[}system{]}$\setminus$wer.dll\end{tabular}}       \\ \hline
\multicolumn{2}{l}{\begin{tabular}[c]{@{}l@{}}WINDOWS\_FILE:Execute:{[}system{]}$\setminus$ntmarta.dll, \\ WINDOWS\_FILE:Execute:{[}system{]}$\setminus$apphelp.dll\end{tabular}} \\ \hline
\multicolumn{2}{l}{\begin{tabular}[c]{@{}l@{}}WINDOWS\_FILE:Execute:{[}system{]}$\setminus$faultrep.dll, \\ WINDOWS\_FILE:Execute:{[}system{]}$\setminus$imm32.dll\end{tabular}}  \\ \hline
\end{tabular}
\caption{Illustration of manipulated features in malware dataset: each feature in this table contains a sequence of two events. The two events in each row happened in the same order as displayed.}
\vspace{-2.5em}
\label{tab:mal_advSample}
\end{table}

\noindent{\textbf{MNIST \& CIFAR-10.}} The MNIST dataset is composed of 70,000 greyscale images (of 28$\times$28, or 784, pixels) of handwritten digits, ranging from 0 to 9. The dataset is split into a training set of 60,000 samples and a test set of 10,000 samples.

The CIFAR-10 dataset consists of 60,000 images, divided into 10 classes. The training split contains 50,000 samples while the test split contains 10,000 samples. Since the samples of CIFAR-10 dataset are color images, each image is made up of 32$\times$32 pixels where each pixel is represented by three color channels (i.e., RGB). 

For the MNIST and CIFAR-10 datasets, we generate adversarial samples by simply adding the adversarial perturbation $\delta{x}$, introduced in Section~\ref{sec:background}), directly to the original image (since feature values are continuous/real-valued). The degree of manipulation is controlled by selecting different $\phi$, as in Equation \eqref{eq:delta_X}.

\subsection{Malware Classification Results}
\label{sec:malware}

\noindent{\textbf{Sensitivity to Nullification Rate}}
We first implement a group of experiments to quantify the effect that nullification rates have on model classification accuracy as well as model resistance. More specifically, we allow the nullification rate to range from 10\% to 90\% with 10\% increments, both at training and testing time. By comparing each experiment result, we may then select the optimal nullification rate. We then integrate our defense mechanism with adversarial training and compare it against all aforementioned methods.

Measures of classification accuracy and model resistance, corresponding to different nullification rates, are shown in Table \ref{tab:1_rate_malware}. As observed in Table~\ref{tab:1_rate_malware}, the classification accuracy of trained models decreases when the nullification rate is increased except when nullification rate is at $40\%$ or $60\%$. These two rates may roughly imply the proportion of noise contained within the original dataset. The average classification accuracy is  $93.66\%$ while the highest achieved is $95.22$, when nullification rate is $10\%$. This shows us that classification performance is more negatively impacted as more important features are discarded. Note that the accuracy remains at a surprisingly high value even when the nullification rate reaches $90\%$. This aligns with the fact that the malware data is quite sparse. 

\begin{table}[t]
\small
\centering
\begin{tabular}{ccc}
\hline
\multirow{2}{*}{\begin{tabular}[c]{@{}c@{}}Expectation of \\ nullification rates (\%)\end{tabular}} & \multicolumn{2}{c}{Malware} \\ \cline{2-3} 
                                    & Accuracy (\%)    & Resistance (\%)    \\ \hline
10                                & 95.22       & 36.46         \\ \hline
20                                & 94.67       & 36.76         \\ \hline
30                                & 93.92       & 38.56         \\ \hline
40                                & 95.20       & 45.19         \\ \hline
50                                & 93.18       & 51.43         \\ \hline
60                                & 93.77       & 49.03         \\ \hline
70                                & 93.10       & 53.96         \\ \hline
80                                & 93.08       & 62.30         \\ \hline
90                                & 90.88       & 64.86         \\ \hline
\end{tabular}
\caption{Classification accuracy vs. model resistance with various feature nullification rates on the malware dataset. Note that the nullification rate hyper-parameter $p$ is simply an expectation, as detailed in Section \ref{sec:tech}, while the other hyper parameter $\sigma$ is set to be $0.05$ for these experiments.}
\vspace{-2.5em}
\label{tab:1_rate_malware}
\end{table}

On the contrary, as shown in Table~\ref{tab:1_rate_malware}, model resistance shows the opposite trend. Maximum resistance against adversarial samples is reached at a 90\% nullification rate. Clearly, with such a high nullification rate, more carefully manipulated features are discarded. The different trends for both classification accuracy and resistance demonstrate well the trade-off between achieving one of the two key goals (i.e., accuracy and robustness). By examining Table~\ref{tab:1_rate_malware}, we adopt 80\% as our feature nullification rate expectation for experiments that follow, as the trained model with this nullification rate maintains the best balance between resistance and accuracy.

\begin{table}[t]
\centering
\begin{tabular}{ccc}
\hline
\multirow{2}{*}{Defense Methods}                          & \multicolumn{2}{c}{Malware} \\ \cline{2-3} 
                                                              & Accuracy (\%)    & Resistance (\%)    \\ \hline
Standard                                                      & 93.99       & 30.00         \\ \hline
Dropout                                                       & 93.16       & 13.96         \\ \hline
Adv Training                                                  & 92.68       & 26.07         \\ \hline
RFN                                                           & 93.08       & 62.30         \\ \hline
\begin{tabular}[c]{@{}c@{}}Adv Training\\ \& RFN\end{tabular} & 94.81       & 68.77         \\ \hline
\end{tabular}
\caption{Classification accuracy vs. model resistance of different learning technologies on the malware dataset. In this table, dropout rates are $50\%$ and feature nullification rates are $80\%$. `Adv Training' simply means adversarial training. Note that 'Standard' means standard deep neural architecture without any regularization.}
\vspace{-2.0em}
\label{tab:2_mal_compare}
\end{table}

\noindent{\textbf{Comparative Results}}
Next, we implement five distinct DNN models by training them with different learning techniques as specified in Table~\ref{tab:2_mal_compare}. We present the architecture of these DNN models as well as the corresponding hyper-parameters in the Appendix. With certain perturbations added to the data samples, Table~\ref{tab:2_mal_compare} first shows that the standard DNN model exhibits poor resistance when classifying adversarial samples. Surprisingly, as shown for dropout and adversarial training, these two methods yield even worse resistance compared to the standard DNN. This strengthens our previous analysis in Section \ref{sec:tech}. Although these mechanisms have been shown to provide certain resistance to already seen adversarial samples and so-called `cross-model' adversarial samples \footnote{Adversarial samples that are crafted from a different DNN that is built to approximate some standard targeted DNN.}, they are even more vulnerable to more specifically crafted adversarial samples. These results are also consistent with those reported in~\cite{gu2014towards}. This implies that the regularization involved in adversarial training and dropout offer poor general resistance to adversarial examples. 

In comparison, RFN provides a significantly better resistance against adversarial samples, as is shown in Table~\ref{tab:2_mal_compare}. The model resistance afforded by our method improves more than $100\%$ (relative error) when comparing with standard DNN. Recall that RFN can also be viewed as a preprocessing approach for the successive DNN. As such, it can be combined with other existing defense mechanisms. It is expected that such a combination would further improve model robustness. In order to verify this, we next combine RNF with adversarial training and compare the hybrid approach to both standalone RFN and adversarial training.

Table~\ref{tab:2_mal_compare} specifies the classification accuracy and model resistance of the hybrid technique. We observe that the combined technique does indeed provide better resistance when compared to standalone RFN. This may due to the fact that RFN and adversarial training penalize adversarial samples in two different manners, and an ensemble of the two favorably amplifies the model resistance that each technique induces. From Table~\ref{tab:2_mal_compare}, we also notice that both standalone RFN and aforementioned combined approach do slightly but noticeably reduce classification accuracy. However, the combination of RFN and adversarial training results in near-negligible degradation. This indicates that RFN, either standalone or when combined with adversarial training, provides much better resistance either adversarial training and dropout on the malware dataset.

\subsection{Image Recognition Results}
In the following experiments, we examine the generality of our proposed method by applying it to the MNIST and CIFAR-10 image recognition tasks. For MNIST, we build a standard feed-forward fully connected DNN, while for CIFAR-10, we build a convolutional neural network (CNN). Similar to the experiments implemented on malware dataset, we also implement two groups of experiments, one for determining the optimal $p$ on each dataset, and another for comparing between different defense technologies. Other hyper-parameters and neural architectural details appear in the Appendix.

\begin{table}[t]
\small
\centering
\begin{tabular}{ccccc}
\hline
\multirow{2}{*}{\begin{tabular}[c]{@{}c@{}}Expectation of \\ nullification rates\end{tabular}} & \multicolumn{2}{c}{MNIST}                                                  & \multicolumn{2}{c}{CIFAR-10}                                                \\ \cline{2-5} 
                                    & Accuracy & \begin{tabular}[c]{@{}c@{}}Resistance\\ $\phi = 0.15$\end{tabular} & Accuracy & \begin{tabular}[c]{@{}c@{}}Resistance\\ $\phi = 0.15$\end{tabular} \\ \hline
10\%                                & 98.17\%         & 70.39\%                                                                & 80.01\%          & 55.87\%                                                                 \\ \hline
20\%                                & 98.09\%         & 73.55\%                                                                & 77.62\%          & 59.55\%                                                                \\ \hline
30\%                                & 97.89\%         & 78.31\%                                                                & 75.95\%          & 61.63\%                                                                \\ \hline
40\%                                & 97.53\%         & 81.49\%                                                                & 74.49\%          & 65.59\%                                                                \\ \hline
50\%                                & 96.78\%         & 83.68\%                                                                & 74.02\%          & 67.85\%                                                                \\ \hline
\end{tabular}
\caption{Classification accuracy vs. model resistance with various feature nullification rates on MNIST and CIFAR-10. Hyper parameter $\sigma$ is also set to be $0.05$ in this evaluation.}
\vspace{-1.2em}
\label{tab:drop_img}
\end{table}

As is shown in Table~\ref{tab:drop_img}, the trend of accuracy and resistance are consistent with that found in the malware experiments. Maximum resistance against adversarial image samples is reached at 50\% nullification rate. With respect to classification accuracy, our proposed method demonstrates roughly similar performance at various nullification rates. Based on this result, we adopted 50\% as our feature nullification rate in the experiments to follow.

In Table~\ref{tab:compare_img}, we show measures of classification accuracy and model resistance of all aforementioned approaches on the MNIST and CIFAR-10 datasets. Much as in the malware experiments, we further evaluate our RFN method combined with adversarial training on both datasets. In Table~\ref{tab:compare_img}, we also measure the resistance of these DNN models against various coefficients $\phi$. 

\begin{table*}[t]
\centering
\begin{tabular}{ccccccccc}
\hline
\multirow{3}{*}{\begin{tabular}[c]{@{}c@{}}Learning \\ Technology\end{tabular}} & \multicolumn{4}{c}{MNIST}                                                 & \multicolumn{4}{c}{CIFAR-10}                              \\ \cline{2-9} 
                                                                                & \multirow{2}{*}{Accuracy} & \multicolumn{3}{c}{Resistance}                & Accuracy & \multicolumn{3}{c}{Resistance}                \\ \cline{3-5} \cline{7-9} 
                                                                                &                           & $\phi = 0.15$ & $\phi = 0.25$ & $\phi = 0.35$ &          & $\phi = 0.15$ & $\phi = 0.25$ & $\phi = 0.35$ \\ \hline
Standard                                                                        & 98.43                     & 8.19          & 0.56          & 0.01          & 73.59    & 19.48         & 13.51         & 10.68         \\ \hline
Dropout                                                                         & 98.61                     & 19.51         & 3.86          & 0.96          & 81.07    & 17.43         & 16.59         & 16.40         \\ \hline
Adv Training                                                                    & 97.46                     & 67.68         & 28.37         & 7.62          & 80.62    & 33.97         & 19.76         & 13.73         \\ \hline
RFN                                                                             & 96.78                     & 83.69         & 71.44         & 60.69         & 74.02    & 67.85         & 51.89         & 41.29         \\ \hline
Adv Training \& RFN                  & 96.11                     & 91.28         & 84.92         & 78.18         & 74.12    & 71.03         & 55.49         & 49.84         \\ \hline
\end{tabular}
\caption{Classification accuracy vs. model resistance with different learning methods, under different $\phi$, for both MNIST and CIFAR-10. In this table, dropout rates and feature nullification rates are set $50\%$ for both datasets.}
\vspace{-1.8em}
\label{tab:compare_img}
\end{table*}

As is shown in Table \ref{tab:compare_img}, adversarial samples generated from a standard DNN are capable of lowering the accuracy of the standard DNN to as low as $0.01\%$ on MNIST and $10.68\%$ on CIFAR-10. In contrast, all of the investigated defense mechanisms yield improved resistance, with, again, models trained with RFN reaching the best level of resistance. In addition, the combination of RFN with adversarial training achieves the best resistance of $91.28$ on MNIST and $74.12\%$ on CIFAR-10. Though different than in the case of malware classification, both dropout and adversarial training alone do provide somewhat improved resistance on both datasets. This indicates that the resistance provided by these methods might be highly dependent on the data type  (images, in this case). In particular, since adversarial training is designed to handle adversarial samples, it demonstrates much better resistance when compared directly to dropout, though both methods offer model regularization.

As for classification accuracy, dropout achieves the highest accuracy on both datasets. For MNIST dataset, both RFN and adversarial training, as well as their combination, do trade some classification accuracy for better resistance. However, for the CIFAR-10 dataset, these methods demonstrate slightly improvement for accuracy. This is due to the fact that the CIFAR-10 task is much more complex than that of MNIST, hence the regularization provided by all of these methods leads to improved generalization. In general, despite the minor accuracy degradation caused by using RFN or the hybrid method, the significant improvement over resistance in both datasets demonstrates that our proposed method is quite promising for classification tasks when resistance to adversarial samples is important. Finally, our method is agnostic to the choice of the DNN architecture, given that we evaluate RFN with both feed-forward fully connected DNNs and CNNs (as evidenced in Table \ref{tab:compare_img}).

\vspace{-0.5em}

\section{Conclusion}
\label{sec:conclusion}

In this paper, we proposed a simple method for constructing deep neural network models that are robust to adversarial samples. Our design is based on a thorough analysis of neural model's vulnerability to adversarial perturbation as well as the limitations of previously proposed defenses. Using our proposed Random Feature Nullification, we have shown that it is impossible for an attacker to craft specifically designed adversarial sample that can force a DNN to misclassify its inputs. This implies that our proposed technology does not suffer, as previous methods do, from attacks that rely on generating model-specific adversarial samples. 

We apply our method to malware dataset and empirically demonstrated that we significantly improve model resistance with only negligible sacrifice of accuracy, compared to other defense mechanisms. Cross-data generality was also demonstrated through experiments in image recognition. Future work will entail investigating the performance of our method to an even wider variety of applications.

\bibliographystyle{ACM-Reference-Format}
\bibliography{ref}

\begin{table*}[h]
\centering
\begin{tabular}{ccccccccc}
\hline
\multicolumn{2}{c}{\multirow{2}{*}{Learning Technologies}} & \multicolumn{7}{c}{Hyper Parameters}                                                            \\ \cline{3-9} 
\multicolumn{2}{c}{}                                     & DNN Structure      & Activation & Optimizer & Learning Rate & Dropout Rate & Batch Size & Epoch \\ \hline
\multicolumn{2}{c}{Standard DNN}                         & 784-784-784-784-10 & Relu       & SGD       & 0.1           &$\times$      & 100        & 25    \\ \hline
\multicolumn{2}{c}{Dropout}                              & 784-784-784-784-10 & Relu       & SGD       & 0.1           & 0.5         & 100        & 25    \\ \hline
\multicolumn{2}{c}{Adv. Training}                        & 784-784-784-784-10 & Relu       & SGD       & 0.01           & 0.5         & 100        & 70    \\ \hline
\multicolumn{2}{c}{RFN}                                  & 784-784-784-784-10 & Relu       & SGD       & 0.1           & 0.25         & 100        & 25    \\ \hline
\multicolumn{2}{c}{RFN \& Adv. Training}                 & 784-784-784-784-10 & Relu       & SGD       & 0.01           & 0.25         & 100        & 70    \\ \hline
\end{tabular}
\caption{The hyper parameters of MNIST models: the network structure specifies the number of hidden layers and hidden units in each layer. The last layer of each model is followed by Softmax non-linearity. Note that standard DNN stands for DNN trained without any regularization.}
\vspace{-1em}
\label{tab:setup_mnist}
\end{table*}

\begin{table*}[h]
\centering
\begin{tabular}{ccccccccc}
\hline
\multicolumn{2}{c}{\multirow{2}{*}{Learning Technology}} & \multicolumn{7}{c}{Hyper Parameters}                                                            \\ \cline{3-9} 
\multicolumn{2}{c}{}                                     & DNN Structure      & Activation & Optimizer & Learning Rate & Dropout Rate & Batch Size & Epoch \\ \hline
\multicolumn{2}{c}{Standard DNN}                         & 5000-1000-100-2     & Relu      & Adam      & 0.001         &$\times$              & 500        & 20    \\ \hline
\multicolumn{2}{c}{Dropout}                              & 5000-1000-100-2      & Relu    & Adam      & 0.001         & 0.5              & 500        & 20    \\ \hline
\multicolumn{2}{c}{Adv. Training}                        & 5000-1000-100-2  & Relu       & SGD      & 0.01           & 0.5         & 500        & 40    \\ \hline
\multicolumn{2}{c}{RFN}                                  & 5000-1000-100-2  & Relu       & Adam       & 0.001           & 0.5         & 500        & 15   \\ \hline
\multicolumn{2}{c}{RFN \& Adv. Training}                 & 5000-1000-100-2  & Relu       & SGD      & 0.01           & 0.5         & 500        & 40    \\ \hline
\end{tabular}
\caption{The hyper parameters of Malware models.}
\vspace{-1em}
\label{tab:setup_malware}
\end{table*}

\begin{table*}[h]
\centering
\begin{tabular}{cccccccc}
\hline
\multicolumn{2}{c}{\multirow{2}{*}{Learning Technology}} & \multicolumn{6}{c}{Hyper parameters}                                       \\ \cline{3-8} 
\multicolumn{2}{c}{}                                     & Activation & Optimizer & Learning rate & Dropout rate & Batch Size & Epoch \\ \hline
\multicolumn{2}{c}{Standard DNN}                         & Relu       & Adam      & 0.001         & $\times$    & 128        & 50    \\ \hline
\multicolumn{2}{c}{Dropout}                              & Relu       & Adam      & 0.001         & 0.5          & 128        & 50    \\ \hline
\multicolumn{2}{c}{Adv. Training}                        & Relu       & SGD       & 0.01          & 0.5          & 128        & 50    \\ \hline
\multicolumn{2}{c}{RFN}                                  & Relu       & Adam      & 0.001         & 0.5          & 128        & 50    \\ \hline
\multicolumn{2}{c}{RFN \& Adv. Training}                 & Relu       & SGD       & 0.01          & 0.5          & 128        & 50    \\ \hline
\end{tabular}
\caption{The hyper parameters of CIFAR-10 models, in this evaluation we use CNN instead of standard DNN}
\vspace{-1em}
\label{tab:hyper_CIFAR}
\end{table*}

\begin{table*}[h]
\centering
\begin{tabular}{cccccc}
\hline
\multirow{2}{*}{Layer type}                     & \multicolumn{5}{c}{Learning Technology}                                \\ \cline{2-6} 
                                                & Standard DNN          & Dropout               & Adv. Training         & RFN                   & RFN \& Adv. Training   \\ \hline
Convolutional                                   & 64 filter($3\times3$) & 64 filter($3\times3$) & 64 filter($3\times3$) & 64 filter($3\times3$) & 64 filter($3\times3$)   \\ \hline
Convolutional                                   & 64 filter($3\times3$) & 64 filter($3\times3$) & 64 filter($3\times3$) & 64 filter($3\times3$) & 64 filter($3\times3$) \\ \hline
Max pooling                                     & $2\times2$            & $2\times2$            & $2\times2$            & $2\times2$            & $2\times2$            \\ \hline
Convolutional                                   & 72 filter($3\times3$) & 72 filter($3\times3$) & 128 filter($3\times3$) & 128 filter($3\times3$) & 128 filter($3\times3$) \\ \hline
Convolutional                                   & 72 filter($3\times3$) & 72 filter($3\times3$) & 128 filter($3\times3$) & 128 filter($3\times3$) & 128 filter($3\times3$) \\ \hline
Max pooling                                     & $2\times2$            & $2\times2$            & $2\times2$            & $2\times2$            & $2\times2$   \\ \hline
Fully Connect                                   & 512 units             & 512 units             & 256 units             & 256 units             & 256 units   \\ \hline
Fully Connect                                   & 256 units             & 256 units             & 256 units             & 256 units             & 256 units \\ \hline
Softmax                                         & 10 units              & 10 units              & 10 units              & 10 units              & 10 units   \\ \hline
\end{tabular}
\caption{The network structure of CIFAR-10 models: the activation function of convolutional layers and fully connected layers are shown in Table \ref{tab:hyper_CIFAR}.}
\vspace{-1em}
\label{tab:struc_CIFAR_1}
\end{table*}

\section{Appendix}

To augment the experimental setup presented in Section~\ref{sec:eval}, the tables in this part contain the hyper parameters used for model training.
\end{document}